\documentclass[11pt]{article}

\usepackage[final]{acl}

\usepackage{times}
\usepackage{latexsym}
\usepackage{booktabs}
\usepackage[T1]{fontenc}
\usepackage[utf8]{inputenc}
\usepackage{microtype}
\usepackage{inconsolata}
\usepackage{graphicx}
\usepackage{amsmath,amssymb}
\usepackage{multirow}
\usepackage{xcolor}
\usepackage{enumitem}
\usepackage{natbib}

\title{When Annotators Agree but Labels Disagree:\\ The Projection Problem in Stance Detection}

\author{Bowen Zhang \\
  Shenzhen Technology University, Shenzhen, China \\
  \texttt{zhang\_bo\_wen@foxmail.com} \\
  }

\begin{document}
\maketitle
\begin{abstract}
Stance detection is nearly always formulated as classifying text into \textit{Favor}, \textit{Against}, or \textit{Neutral}. This convention was inherited from debate analysis and has been applied without modification to social media since SemEval-2016. However, attitudes toward complex targets are not unitary. A person can accept climate science while opposing carbon taxes, expressing support on one dimension and opposition on another. When annotators must compress such multi-dimensional attitudes into a single label, different annotators may weight different dimensions, producing disagreement that reflects different compression choices rather than confusion. We call this the \textbf{projection problem}. We conduct an annotation study across five targets from three stance benchmarks (SemEval-2016, P-Stance, COVID-19-Stance), with the same three annotators labeling all targets. For each target, annotators assign both a standard stance label and per-dimension judgments along target-specific dimensions discovered through bottom-up analysis, using the same number of categories for both. Across all fifteen target--dimension pairs, dimensional agreement consistently exceeds label agreement. The gap appears to scale with target complexity: modest for a single-entity target like Joe Biden (AC1: 0.87 vs.\ 0.95), but large for a multi-faceted policy target like school closures (AC1: 0.21 vs.\ 0.71).
\end{abstract}

\section{Introduction}
\label{sec:intro}

After nearly a decade of research, stance detection appears to be approaching maturity. On the SemEval-2016 benchmark \cite{mohammad2016semeval}, $F_\text{avg}$ has climbed from below 60 with early classifiers to above 80 with LLM-based pipelines \cite{zhang2026induce}. New methods incorporate background knowledge, multi-agent reasoning, and chain-of-thought prompting \cite{lan2024cola, zhang2025logic, dai2025large}, pushing numbers higher with each iteration.

Yet all of this progress rests on a shared assumption: that a person's stance toward a target can be adequately captured by a single label from \{Favor, Against, Neither\}. Consider what happens when that assumption meets a real text. The target is \textit{Climate Change Is Real Concern}:

\begin{quote}
\small\textit{``Climate change is real, but carbon tax will destroy the economy and hurt working families.''}
\end{quote}

\noindent Ask three annotators to label this text. One labels it \textsc{Favor}, because the author accepts that climate change is real. Two label it \textsc{Against}, because the author opposes the primary policy response. The majority vote yields \textsc{Against}. The dissenter is recorded as having made an error.

Did they? Both sides understood the text perfectly well. Their disagreement was not about what the text \textit{says} but about which part of what it says \textit{matters more} for the overall label. The text affirms a factual claim and opposes a policy. These are two signals pointing in opposite directions, and the label scheme forces a choice between them.

This paper argues that such cases are not annotation failures but a structural consequence of the task definition. The three-way taxonomy was originally developed for debate analysis, including congressional speeches \cite{thomas2006get} and online forums \cite{somasundaran2009recognizing}, where speakers explicitly take sides and a single label is natural. When this framework was transplanted to social media \cite{mohammad2016semeval}, it carried over an assumption that often fails: that attitudes come in one flavor at a time. We call this the \textbf{projection problem}. Annotators perceive stance as multi-dimensional but must compress it into a single category. Different annotators compress differently, and disagreement follows. We do not claim that three-way labels are wrong, but rather that their representational limitations, not model limitations, may be becoming a binding constraint on progress.

\section{Attitudes Are Multi-Dimensional}
\label{sec:multi}

A person's position on \textit{Climate Change Is Real Concern} can vary independently along multiple dimensions. Consider three: acceptance of the science, support for policy responses, and urgency of action. These are logically independent. One can accept the science while opposing carbon taxes, or acknowledge urgency while rejecting specific legislation. Table~\ref{tab:tweets} shows five tweets and their approximate dimensional profiles, using the same three-way judgment (Support / Against / not addressed) that our annotation study employs.

\begin{table*}[t]
\centering
\small
\begin{tabular}{p{6.4cm}cccccc}
\toprule
\textbf{Tweet} & \textbf{D1} & \textbf{D2} & \textbf{D3} & \textbf{Label} & \textbf{Disagree}\\
& \footnotesize Science & \footnotesize Policy & \footnotesize Urgency & & \footnotesize risk\\
\midrule
(a) \textit{``Global warming is a HOAX!''} & Ag. & -- & -- & Ag. & Low \\[2pt]
(b) \textit{``Fix poverty first. Climate can wait.''} & -- & -- & Ag. & Ag. & Low \\[2pt]
(c) \textit{``We must act NOW. Our children deserve a livable planet.''} & Sup. & Sup. & Sup. & Fav. & Low \\[2pt]
(d) \textit{``Climate change is real, but carbon tax will destroy the economy.''} & Sup. & Ag. & -- & \textbf{???} & \textbf{High} \\[2pt]
(e) \textit{``ONE Volcano emits more pollution than man has in our HISTORY!''} & Ag. & -- & -- & Ag. & Low \\
\bottomrule
\end{tabular}
\caption{Five CC tweets with approximate dimensional profiles (``--'' = not addressed). For tweets (a)--(c) and (e), all addressed dimensions agree in direction, so any weighting yields the same label. For tweet (d), dimensions conflict: Science is supportive while Policy is opposing. The overall label depends on which dimension the annotator prioritizes.}
\label{tab:tweets}
\end{table*}

The key observation is that tweets (a)--(c) and (e) have consistent dimensions. Any annotator arrives at the same label regardless of weighting. Tweet (d) has conflicting dimensions, so the label depends on which dimension the annotator prioritizes. Note also that (a) and (b) are both Against but structurally different: (a) reflects science denial, while (b) reflects deprioritization. Three-way labels conflate these distinct stances.

The idea that attitudes are multi-dimensional is well established in psychology. Multi-item scales are standard precisely because single-item measures lose information \cite{devellis2021scale, krosnick1999survey}. NLP's three-way label is, in social science terms, a single-item categorical measure applied to a multi-dimensional construct.

\section{The Projection Problem}
\label{sec:projection}

When an annotator assigns a three-way label, the process can be understood as involving two steps. The first step is \textbf{perception}: the annotator reads the text and forms a judgment on each relevant dimension. The second step is \textbf{projection}: the annotator compresses these per-dimension judgments into a single label by implicitly weighting dimensions. Different annotators may perceive similar signals in the first step but weight them differently in the second step, producing different labels.

This two-step account is consistent with several documented phenomena. In the SEM16 annotation, even among tweets that passed quality filtering, pairwise agreement was only 73\%, and roughly one in four tweets were discarded for failing to reach 60\% majority \cite{mohammad2016dataset}. The annotation guidelines themselves acknowledge that a tweeter may ``support the target to some extent, but [be] also against it to some extent,'' yet this case was absorbed into a ``Neutral'' category accounting for less than 0.1\% of labels.

\subsection{Predictions}
\label{sec:predictions}

If disagreement on stance labels is driven primarily by the projection step rather than the perception step, two predictions follow.

\paragraph{Prediction 1 (Global).} For any target, annotator agreement on individual dimensions should tend to exceed agreement on the overall label. This is because dimensional annotation bypasses the projection step. Annotators judge each dimension directly, without needing to compress multiple signals into one category.

\paragraph{Prediction 2 (Gradient).} The gap between dimensional and label agreement should vary across targets. Some targets may be dominated by a single dimension, where different weightings converge on the same label and projection is nearly lossless. Other targets may involve multiple weakly correlated dimensions, where different weightings yield different labels and the gap should be larger.

\section{Annotation Study}
\label{sec:study}

\subsection{Setup}

\paragraph{Targets and data.} We select five targets spanning three widely used stance detection benchmarks:

\begin{itemize}[leftmargin=*,itemsep=1pt]
\item \textbf{SemEval-2016 Task 6} \cite{mohammad2016semeval}: \textit{Climate Change is a Real Concern} (CC) and \textit{Legalization of Abortion} (LA).
\item \textbf{P-Stance} \cite{li2021pstance}: \textit{Joe Biden} (Biden).
\item \textbf{COVID-19-Stance} \cite{glandt2021stance}: \textit{Anthony S.\ Fauci} (Fauci) and \textit{School Closures} (SC).
\end{itemize}

\noindent These targets cover a range of complexity, from single-entity targets to multi-faceted policy targets. For each target, we randomly sample 60 texts from the original dataset, yielding 300 texts in total.

\paragraph{Annotators.} Three graduate students in NLP, fluent in English, independently labeled all five targets. Using the same annotators throughout creates a within-subject comparison, so that differences in agreement across targets are less likely to reflect differences in annotator quality.

\paragraph{Dimension discovery.} Rather than imposing a universal set of dimensions across all targets, we identify target-specific dimensions through a semi-automated bottom-up process. A large language model is first used to analyze the sampled texts for each target and extract recurring attitudinal dimensions. The resulting candidates are then reviewed, consolidated, and finalized by the research team. This process yields three dimensions per target (Table~\ref{tab:dimensions}).

\begin{table}[t]
\centering
\small
\begin{tabular}{ll}
\toprule
\textbf{Target} & \textbf{Dimensions} \\
\midrule
\multirow{3}{*}{CC} & D1: Impact on Biodiversity \\
& D2: Government and Policy \\
& D3: Public Awareness \& Activism \\
\midrule
\multirow{3}{*}{LA} & D1: Human Life Debate \\
& D2: Bodily Autonomy \\
& D3: Moral and Religious Views \\
\midrule
\multirow{3}{*}{Biden} & D1: Electoral Momentum \\
& D2: Health \& Competence Concerns \\
& D3: Policy Perspectives \\
\midrule
\multirow{3}{*}{Fauci} & D1: Public Health Messaging \\
& D2: Trust and Credibility \\
& D3: Political Polarization \\
\midrule
\multirow{3}{*}{SC} & D1: Health and Safety Concerns \\
& D2: Equity and Access Issues \\
& D3: Gov.\ and Policy Critique \\
\bottomrule
\end{tabular}
\caption{Target-specific dimensions identified through bottom-up analysis.}
\label{tab:dimensions}
\end{table}

\paragraph{Annotation scheme.} For each text, annotators provided two layers of judgment: (1) a standard \textbf{stance label}, and (2) a \textbf{per-dimension stance judgment} for each of three dimensions, with an explicit N/A option for dimensions the text does not address. Both layers use the same number of categories for a given dataset: Favor / Against for SEM16 and P-Stance targets, and Favor / Against / Neutral for COVID-19-Stance targets. This ensures that any difference in agreement between labels and dimensions is not attributable to differences in the number of response options.

\paragraph{Agreement metrics.} We report two metrics: \textbf{observed agreement} (the proportion of annotator pairs assigning the same category over non-N/A annotations) and \textbf{Gwet's AC1} \cite{gwet2008computing}, a chance-corrected coefficient. We choose AC1 over Krippendorff's $\alpha$ or Cohen's $\kappa$ because AC1 is more robust to the prevalence paradox, where highly skewed category distributions cause $\alpha$ and $\kappa$ to yield paradoxically low values even when raw agreement is high \cite{zhao2013assumptions, zhang2022sentiment,NiuMM,ding-etal-2024-edda,10094597,10537616,DAI2025106956,Sun22}.

\subsection{Results}
\label{sec:results}

Table~\ref{tab:main} presents the core results.

\begin{table}[t]
\centering
\small
\begin{tabular}{l l ccccc}
\toprule
& \textbf{Target} & \textbf{Label} & \textbf{D1} & \textbf{D2} & \textbf{D3} & \textbf{Dim avg} \\
\midrule
\multirow{5}{*}{\rotatebox{90}{\footnotesize Agreement}}
& CC & .708 & .875 & .818 & .917 & .870 \\
& LA  & .893 & .966 & .903 & .913 & .927 \\
& Biden & .932 & .959 & .971 & .958 & .963 \\
& Fauci & .706 & .905 & .871 & .759 & .845 \\
& SC & .472 & .802 & .857 & .667 & .775 \\
\midrule
\multirow{5}{*}{\rotatebox{90}{\footnotesize AC1}}
& CC & .559 & .859 & .771 & .904 & .845 \\
& LA & .811 & .961 & .888 & .903 & .917 \\
& Biden & .870 & .946 & .965 & .946 & .952 \\
& Fauci & .584 & .877 & .837 & .684 & .799 \\
& SC  & .214 & .749 & .809 & .564 & .707 \\
\bottomrule
\end{tabular}
\caption{Observed agreement and Gwet's AC1 for five targets ($n{=}60$ each, 3 annotators). \textbf{Label}: overall stance label. \textbf{D1--D3}: individual dimensions (see Table~\ref{tab:dimensions}). \textbf{Dim avg}: mean of D1--D3. Across all targets and dimensions, dimensional agreement exceeds label agreement under both metrics.}
\label{tab:main}
\end{table}

\paragraph{Prediction 1 supported.}

All 15 individual target--dimension pairs show dimensional AC1 higher than the corresponding label AC1 (binomial test, $p < 0.001$). The same holds for observed agreement (15/15). The pattern also holds under a stricter test: for every target, even the weakest individual dimension exceeds the label AC1 (Table~\ref{tab:delta}). This suggests that the dimensional advantage is not an artifact of averaging across a mix of strong and weak dimensions.

\paragraph{Prediction 2 supported.}

We define the projection loss $\Delta$ as the difference between dimensional average AC1 and label AC1. We additionally report a retention rate, defined as the ratio of label AC1 to dimensional average AC1, capturing how much of the achievable agreement survives projection. Table~\ref{tab:delta} ranks the five targets by $\Delta$.

\begin{table}[t]
\centering
\small
\begin{tabular}{lccccc}
\toprule
\textbf{Target} & \textbf{Label} & \textbf{Dim} & \textbf{Min} & $\boldsymbol{\Delta}$ & \textbf{Ret.} \\
& \footnotesize AC1 & \footnotesize avg & \footnotesize dim & & \\
\midrule
SC     & .214 & .707 & .564 & \textbf{.493} & 30\% \\
CC     & .559 & .845 & .771 & .286 & 66\% \\
Fauci  & .584 & .799 & .684 & .215 & 73\% \\
LA     & .811 & .917 & .888 & .106 & 88\% \\
Biden  & .870 & .952 & .946 & .082 & 91\% \\
\bottomrule
\end{tabular}
\caption{Projection loss $\Delta = \text{Dim avg AC1} - \text{Label AC1}$ and retention rate (Label AC1 / Dim avg AC1), ranked by $\Delta$. \textbf{Min dim}: the lowest-agreement individual dimension for each target.}
\label{tab:delta}
\end{table}

The gradient is broadly consistent with expectations about target complexity. School Closures, a policy target involving health, education equity, and governance, shows the largest projection loss ($\Delta = 0.49$, 30\% retention). Biden shows the smallest ($\Delta = 0.08$, 91\% retention), consistent with the expectation that political alignment dominates this target. Because the same three annotators labeled all five targets, the gradient is unlikely to reflect differences in annotator quality.

This gradient is also consistent with indirect evidence from prior work. \citet{zhang2026msme} report that MSME's Label Expert, which decomposes labels into sub-categories, yields $+9.4$ F1 on Climate Change but only $+1.7$ on Donald Trump. A method that compensates for label coarseness helps most where labels are likely coarsest.

\paragraph{Dimension quality varies with concreteness.}

Not all dimensions are equally easy to judge. The highest-agreement dimensions tend to be those with concrete lexical cues, such as \textit{Human Life Debate} on Abortion (AC1 = 0.961) and \textit{Health \& Competence Concerns} on Biden (0.965). The lowest-agreement dimensions tend to be more abstract, such as \textit{Political Polarization} on Fauci (0.684) and \textit{Government and Policy Critique} on School Closures (0.564). This suggests that broadly scoped dimensions can themselves suffer a mild form of the projection problem, and that future work on dimensional annotation may benefit from prioritizing concrete, operationalizable dimensions.

\paragraph{Implications for benchmark construction.}

When SEM16 was constructed, roughly 25\% of tweets were discarded because annotators could not reach 60\% majority \cite{mohammad2016dataset}. Our data suggest that these discarded tweets may have been disproportionately dimension-conflicting cases. The published benchmark, by filtering them out, may over-represent texts where three-way labels happen to work well.

\section{Implications}
\label{sec:implications}

\paragraph{Evaluation.} A model trained on majority-voted labels learns to replicate the majority's projection function. On dimension-consistent texts, which likely constitute the majority of current benchmarks, this may suffice. On dimension-conflicting texts, the model must match what may be an arbitrary tie-breaking preference. Rising F1 could therefore partly reflect improved projection mimicry rather than improved stance understanding. Our findings suggest several complements to standard F1: conflict-stratified evaluation that reports performance separately on dimension-consistent and dimension-conflicting texts; dimension-level metrics that measure per-dimension agreement between model predictions and human judgments; and disagreement-aware evaluation against the full annotator distribution rather than the majority label \cite{uma2021learning, davani2021dealing}.

\paragraph{Benchmark augmentation.} Our study suggests that multi-dimensional annotation is feasible and can recover agreement that labels lose. A two-layer scheme providing standard stance labels alongside per-dimension judgments with N/A would preserve backward compatibility while adding dimensional structure. The annotation design cost is manageable through semi-automated dimension discovery, and on complex targets the payoff is substantial: dimensional annotation recovers usable agreement where three-way labels provide little signal.

\section{Conclusion}

Across five targets from three benchmarks, labeled by the same three annotators, dimensional agreement exceeds label agreement in all 15 target--dimension pairs ($p < 0.001$). The projection loss scales with target complexity, from $\Delta = 0.08$ on Biden (91\% retention) to $\Delta = 0.49$ on School Closures (30\% retention). These results suggest that annotators perceive the dimensional structure of texts with considerable consistency, and that disagreement on stance labels may largely reside in the projection step rather than in the perception step. We suggest that the field consider augmenting standard annotation with dimensional judgments and developing evaluation methods that can distinguish projection mimicry from attitude understanding.

\section*{Limitations}

The study uses 300 texts across five targets with three annotators. While the within-subject design and consistency across all 15 target--dimension pairs provide encouraging evidence, a larger annotator pool would yield more stable agreement estimates. Dimensions were discovered through LLM-assisted bottom-up analysis and finalized by the research team. While this is more principled than purely top-down definition, it still involves researcher judgment, and alternative dimension sets might yield different agreement patterns. Some dimensions showed weaker agreement, suggesting that the quality of dimensional annotation depends on how well dimensions are defined. Finally, the projection problem appears most acute for complex, multi-faceted targets. Simpler targets show small projection loss and may be adequately served by three-way labels.

\section*{Ethics Statement}

Annotators participated with informed consent and were compensated at standard local rates. Our advocacy for preserving minority annotator perspectives rather than suppressing them via majority voting has positive implications for representing diverse viewpoints on contested social issues.

\bibliography{custom}

\end{document}